\documentclass{article}

\usepackage{microtype}
\usepackage{graphicx}
\usepackage{subfigure}
\usepackage{booktabs} 

\usepackage{algorithm,algorithmic}

\usepackage{hyperref}

\usepackage[accepted]{icml2025}

\usepackage{amsmath}
\usepackage{amssymb}
\usepackage{mathtools}
\usepackage{amsthm}

\usepackage{multirow}
\usepackage{multicol}
\usepackage{makecell}

\usepackage[capitalize,noabbrev]{cleveref}

\theoremstyle{plain}

\theoremstyle{definition}

\theoremstyle{remark}

\usepackage[textsize=tiny]{todonotes}

\begin{document}

\twocolumn[
\icmltitle{Ultra-Low-Latency Spiking Neural Networks with Temporal-Dependent Integrate-and-Fire Neuron Model for Objects Detection}




\begin{icmlauthorlist}
\icmlauthor{Chengjun Zhang}{wioe}
\icmlauthor{Yuhao Zhang}{wioe}
\icmlauthor{Jie Yang}{wlu}
\icmlauthor{Mohamad Sawan}{wioe,wlu}
\vskip 0.3in
\end{icmlauthorlist}

\icmlaffiliation{wioe}{Zhejiang Key Laboratory of 3D Micro/Nano Fabrication and Characterization, Westlake Institute for Optoelectronics, Fuyang, Hangzhou 311421, China}
\icmlaffiliation{wlu}{CenBRAIN Neurotech, School of Engineering, Westlake University, Hangzhou, 310030, Zhejiang, China}

\icmlcorrespondingauthor{Jie Yang}{yangjie@westlake.edu.cn}
\icmlcorrespondingauthor{Mohamad Sawan}{sawan@westlake.edu.cn}

]



\printAffiliationsAndNotice{}  

\begin{abstract}
Spiking Neural Networks (SNNs), inspired by the brain, are characterized by minimal power consumption and swift inference capabilities on neuromorphic hardware, and have been widely applied to various visual perception tasks.
Current ANN-SNN conversion methods have achieved excellent results in classification tasks with ultra-low time-steps, but their performance in visual detection tasks remains suboptimal.
In this paper, we propose a delay-spike approach to mitigate the issue of residual membrane potential caused by heterogeneous spiking patterns.
Furthermore, we propose a novel temporal-dependent Integrate-and-Fire (tdIF) neuron architecture for SNNs. 
This enables Integrate-and-fire (IF) neurons to dynamically adjust their accumulation and firing behaviors based on the temporal order of time-steps.
Our method enables spikes to exhibit distinct temporal properties, rather than relying solely on frequency-based representations.
Moreover, the tdIF neuron maintains energy consumption on par with traditional IF neuron.
We demonstrate that our method achieves more precise feature representation with lower time-steps, enabling high performance and ultra-low latency in visual detection tasks.
In this study, we conduct extensive evaluation of the tdIF method across two critical vision tasks: object detection and lane line detection.
The results demonstrate that the proposed method surpasses current ANN-SNN conversion approaches, achieving state-of-the-art performance with ultra-low latency (within 5 time-steps). Our code is publicly available at \href{https://github.com/zhangcj13/tdIF}{https://github.com/zhangcj13/tdIF}.
\end{abstract}

\section{Introduction}
\label{submission}
Inspired by the brain's neuronal dynamics \cite{roy2019towards}, Spiking Neural Networks (SNNs) \cite{maass1997networks}  are anticipated to offer a low-power alternative to Artificial Neural Networks (ANNs), leveraging sparse, event-driven spikes for information transmission and processing.
Over the past decade, ANNs have garnered significant attention across various industries and have been widely applied in fields such as computer vision, natural language processing, robotic control, etc.
However, the increasing complexity of tasks and datasets necessitates more complex ANN architectures with larger parameter counts, which results in the increased demands for computational resources and energy consumption \cite{hubara2016binarized}.
At this point, the low-energy consumption advantage of SNNs becomes evident as neurons in SNNs remain inactive and consume no energy until they receive a spike/event afferent \cite{christensen20222022}.
From the perspective of computational hardware operations, multiplication operations in state-of-the-art deep ANNs are replaced by accumulate (AC) operations in SNNs, resulting in significantly lower energy consumption.
In addition, SNNs enable a more power-efficient event-driven computing paradigm on neuromorphic hardware through the substitution of dense multiplication with sparse addition \cite{han2020rmp}.
Meanwhile, as demonstrated in \cite{merolla2014million,davies2021advancing,painkras2013spinnaker}, SNNs implemented on designated neuromorphic processors can achieve energy consumption and latency that are several orders of magnitude lower than ANNs.
Within the realm of neuromorphic computing, scholars are advancing neuromorphic computing architectures tailored for SNN applications, exemplified by systems such as TrueNorth \cite{merolla2014million}, Loihi \cite{davies2018loihi} and Darwin3 \cite{ma2024darwin3}.
Additionally, numerous specialized hardware accelerators \cite{2023C, fang2024energy} tailored for SNNs have been implemented to optimally utilize spike-driven computation properties.
These architectures, innovatively structured to mitigate the von Neumann bottleneck through the integration of memory and computation units, facilitate energy-efficient SNN inference processes.

Currently, two mainstream methodologies exist for developing supervised deep SNNs: direct training and conversion from ANNs. 
Gradient-descent backpropagation \cite{rumelhart1986learning}, which has been highly successful in ANNs, can not be directly applicable to SNNs due to the discontinuous functionality of spiking neurons.
To address the non-differentiability issue of spiking functions, recent researchs have achieved substantial advancements in the direct training of SNNs through the application of backpropagation algorithms, facilitated by the utilization of surrogate gradients \cite{wu2018spatio,gu2019stca,fang2021incorporating}.
Subsequently, they employ Backpropagation Through Time (BPTT) to optimize SNNs, which mirrors the backpropagation technique employed in conventional ANNs.
However, computational and memory inefficiencies persist in this method, especially when implemented on common GPU platforms \cite{wu2021progressive}, owing to the absence of tailored optimizations for binary event handling.
Additionally, the sparsity of spike trains and the vanishing/exploding gradient problem further limit the effectiveness of direct training SNNs for high-complexity tasks.

In contrast, ANN-SNN conversion approaches, as an alternative solution, employ identical training paradigms as ANNs, thereby inheriting their computational efficiency advantages while requiring significantly fewer training resources than direct SNN training techniques \cite{cao2015spiking,diehl2015fast}. 
The primary goal of ANN-SNN conversion is to align analog neuron activations with spiking neuron outputs, typically measured via firing rates \cite{diehl2015fast}.
Through emulating ANN activation patterns via SNN firing rates, the ANN-SNN conversion paradigm has achieved remarkable accuracy in challenging applications, attaining performance levels comparable to conventional ANNs \cite{hu2021spiking,kim2020spiking,yan2021near}.
However, all existing methods suffer from quantization and residual potential errors (as detailed later in Section \ref{sec:cvt_err}), resulting in compromised performance metrics throughout the conversion procedure, particularly in cases of low-latency in visual detection tasks which require precise representations.
While increasing inference latency mitigates these errors, it incurs higher computational costs and delays, undermining SNNs’ efficiency advantages.

In this work, we implement a low-latency and high-performance ANN-SNN conversion method for visual detection tasks. Our main contributions are summarized as follows:
\begin{itemize}

\item Through mathematical analysis, we characterize the operational equivalence between ANN and SNN computational units, and focus on analyzing  residual potential errors (RPE) and quantization errors (QE).
\item We propose delay-spike strategy to minimize RPE and develop a dedicated inference pipeline to minimize the operational delays associated with delay-spike approaches in actual inference.
\item We develop a tdIF (temporal-dependent Integrate-and-Fire Neuron) model that enables neurons to effectively utilize temporal information, thereby significantly reducing the latency of SNNs.
Meanwhile, SNN models utilizing tdIF neurons demonstrate energy consumption levels within the same order of magnitude as those using IF neurons during inference.
\item We validate the effectiveness of our algorithm on two visual detection tasks: object detection (at 5 time-steps, achieving mAP of $74.41\%$  on PASCAL VOC, $55.73\%$ on COCO) and lane line detection (at 5 time-steps , achieving Acc of $95.48\%$ on Tusimple, F1 of $76.65\%$ on CULane)
\end{itemize}

\section{Related Work}
\label{sec:rlw}
We will review current research progress from two key perspectives: ANN-SNN conversion methods and SNN-based detection methods.

\subsection{ANN-SNN Conversion}

The seminal work on ANN-SNN conversion by \cite{cao2015spiking} established the functional correspondence between rectified linear unit (ReLU) activation and integrate-and-fire (IF) neuronal dynamics.
The weight scaling method based on maximum possible activations  is proposed by Diehl et al. \cite{diehl2015fast} to address quantization errors, which are identified as the key impediment to lossless ANN-SNN conversion.
Rueckauer et al. \cite{rueckauer2017conversion} enhance the weight normalization technique by employing the $99.9th$ percentile of neuronal activations.
Additionally, they introduce the “reset-by-subtraction” mechanism, also referred to as “soft-reset” mechanism, in order to mitigate the requirement for excessive time-steps.
Meanwhile, Han et al. \cite{han2020rmp} employ soft-reset spiking neurons with residual membrane potential (RMP) retention to better approximate ReLU functionality and mitigate ANN-SNN conversion loss.
To eliminate conversion errors, many recent approaches \cite{deng2021optimal, ding2021optimal} have adopted the strategy of replacing the activation functions in source ANN training.
A Rate Normalization Layer is developed by Ding et al. \cite{ding2021optimal} to replace ReLU activation functions in source ANN training, thereby enabling seamless conversion to SNNs.
Bu et al. \cite{bu2023optimal} introduce a novel quantization clip-floor-shift (QCFS) activation mechanism for ANN training, specifically designed to reduce conversion errors and enable high-accuracy SNN implementation with minimal time-steps.
Jiang et al. \cite{jiang2023unified} present SlipReLU, a hybrid activation function combining threshold-ReLU and step function characteristics through weighted summation, effectively balancing the dual objectives of maintaining ANN accuracy while minimizing SNN conversion errors.
However, these methods focus exclusively on classification tasks and lack comprehensive experimental validation for visual detection tasks.

\subsection{Detection based on SNNs}
For detection tasks, recent advancements in ANN-SNN conversion have enabled the conversion of pre-trained ANN detectors into SNN implementations.
Notably, Spiking-YOLO \cite{kim2020spiking} achieves performance comparable to conventional ANN detectors through specialized  norm techniques, though demanding $8000$ time-steps to match its ANN equivalent's accuracy. 
Farhadi et al. \cite{li2022spike} utilize Spike Calibration (SpikCalib) to detect SIN errors and rectify false spikes, enabling the accomplishment of various visual perception tasks, while still take $512$ time-steps to achieve good performance.
Wang et al.'s novel two-stage conversion algorithm \cite{wang2023toward} significantly minimized three critical error sources in ANN-SNN conversion: quantization errors, clipping errors, and residual membrane potential errors, while maintaining detection capability with $40$ time-steps.
Through optimizing the clipping threshold and designing layer-specific distribution patterns for synaptic weights and neuronal activations, Fast-SNN \cite{hu2023fast} accomplishes remarkable object detection accuracy with an exceptionally low-latency of $7$ time-steps.
Nevertheless, these low-latency ANN-SNN conversion methods still require large time-steps \cite{kim2020spiking, li2022spike} or additional training costs after conversion \cite{wang2023toward,hu2023fast} to get better performance on detection tasks.

Meanwhile, recent studies have explored the implementation of surrogate gradient methods \cite{wu2018spatio} in detection tasks.
Using surrogate gradient learning, Cordone et al. \cite{cordone2022object} integrate a SNN into an SSD framework for object detection purposes.
Su et al. \cite{su2023deep} introduce a pioneering framework, EMS-YOLO, enabling direct training of SNNs for object detection tasks and achieving competitive accuracy within merely $4$ time-steps.
Yuan et al. \cite{yuan2024trainable} develop an end-to-end object detection SNN framework leveraging a surrogate gradient-enhanced BPTT algorithm, achieving competitive performance within $5$ time-steps.
Luo et al. \cite{luo2024integer} propose the SpikeYOLO framework integrating an I-LIF spiking neuron design, enabling integer-domain training protocols and event-triggered inference to push the performance limits of SNN.
Direct training SNN methods, though capable of operating with reduced time-steps, are constrained by gradient instability (vanishing/exploding), resulting in inferior performance compared to ANN.
In contrast to ANNs, the direct training of SNNs is notably more resource-intensive, requiring both increased memory allocation and roughly T times the computational duration.

For the above reasons, our approach prioritizes a one-stage ANN-SNN conversion framework for detection tasks.
Our work shares similarities with the work presented in \cite{bu2023optimal,deng2021optimal}, which focuses on optimal conversion. 
However, our approach simultaneously reduces quantization and residual membrane potential errors, enabling visual detection at lower time-steps.

\section{Preliminaries}
\label{submission}
\textbf{Analog Neuron Model.}
In the context of ANNs, the computations performed by analog neurons can be streamlined as a concatenation of a linear transformation and a subsequent non-linear mapping. Specifically,  analog neuron is formulated as detailed subsequently:
\begin{align}
        \label{Eq:activition_base} a^{l} &=h(\mathbf{W}^{l}a^{l-1})
\end{align}
where $l=1,\dots,L$ is the index of each layer, $a^{l}$ is the output of the ReLU activation function $h(\cdot)$, and $\mathbf{W}^{l}$ represents the weight matrix connecting layer $l$ to layer $l-1$.

\textbf{Spiking Neuron Model.}
IF neuron model, which is characterized by neuronal dynamics that enable spatiotemporal information processing and provides a robust representation of firing rate, is often employed in conversion algorithms for SNNs \cite{diehl2015fast,rueckauer2017conversion,sengupta2019going}.
The dynamics of IF neurons can be characterized:
\begin{align}
        \label{Eq:charge_func} M^{l}[t] &=V^{l}[t-1] + \mathbf{W}^{l}S^{l-1}[t]\\
        \label{Eq:fire_func} S^{l}[t] &=\Theta(M^{l}[t]-\theta^{l})\\
        \label{Eq:reset_func} V^{l}[t] &=M^{l}[t]-\theta^{l}S^{l}[t]
\end{align}
where $t$ is the time-step of SNNs, $M^{l}[t]$ captures the subthreshold membrane potential before spike initiation, while $V^{l}[t]$ reflects the potential after spike emission, $\theta^{l}$ is the firing threshold.
The IF neuron generates spikes $S^{l}[t]$ with $\Theta(x)$ which represents the Heaviside step function, where $\Theta(x)=1$  when $x>=0$ , and $\Theta(x)=0$ when $x<0$.
The soft-reset (reset-by subtraction) mechanism \cite{han2020rmp,cao2015spiking} which is presented as Eq. \ref{Eq:reset_func} 
is adopted to minimize the loss of information when the membrane potential surpasses the threshold, instead of resetting it to a constant value.

\section{ANN-SNN Conversion Error}\label{sec:cvt_err}

\begin{figure*}[t]
	\begin{center}
		\includegraphics[width=1.0\linewidth]{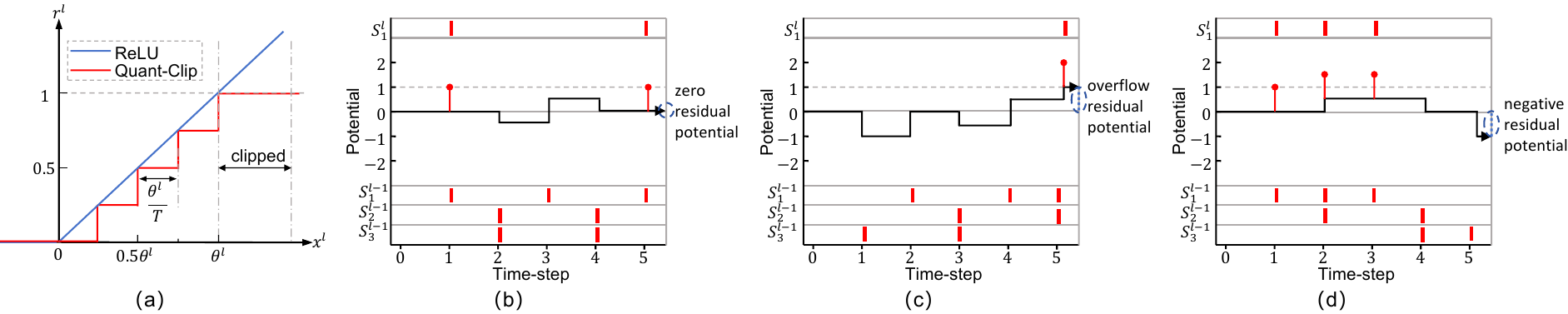}
	\end{center}
    \vspace{-0.4cm}
	\caption{Conversion error between source ANN and converted SNN. (a) Errors between ReLU and Quant-Clip controlled by threshold $\theta^l$ and time-steps $T$. (b$\sim$d) Handcrafted examples of residual membrane potential representation error under the case of (b) zero residual potential, (c) overflow residual potential and (d) negative residual potential.}
	\label{fig:smm_and_rl}
\end{figure*}

\begin{figure}[t]
	\begin{center}
		\includegraphics[width=1.0\linewidth]{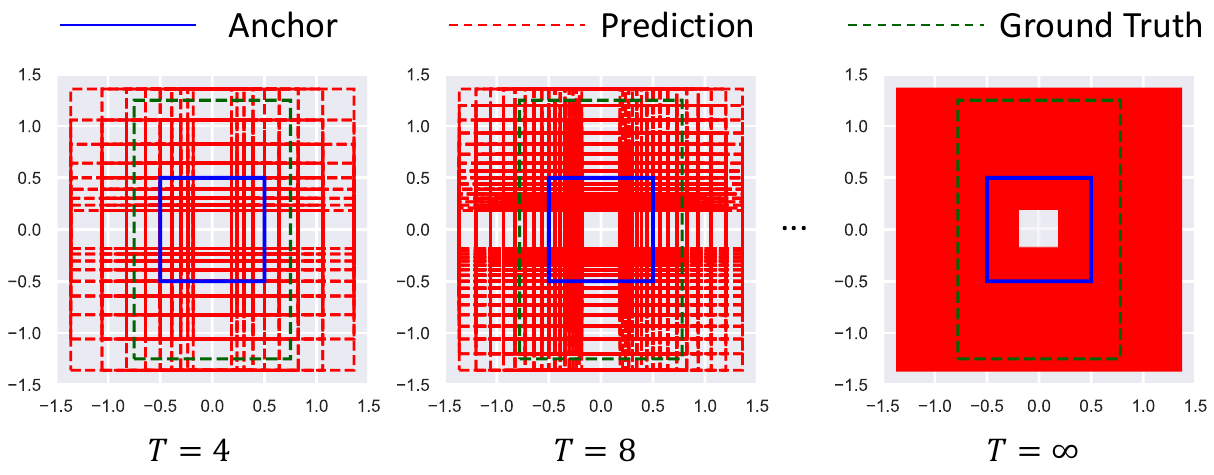}
	\end{center}
    \vspace{-0.4cm}
	\caption{
        Illustration of inaccuracy in matching due to quantization errors at low time-steps in simplified bounding box regression based on YOLO anchor.
    }
	\label{fig:quant_err}
\end{figure}

The objective of ANN-SNN conversion aims to create a consistent relationship where the spiking rate $r^{l}$ in SNNs represents the analog activation $a^l$ in ANNs.
From the dynamics described in Eqs. \ref{Eq:charge_func}\textasciitilde  \ref{Eq:reset_func}, we can derive the functional dependence of membrane potential $V^l$ on the input firing rate $r^{l-1}$ as:
\begin{align}
        \label{Eq:sr_eq} \frac{V^l[T]-V^l[0]}{T} &= \mathbf{W}^{l} \sum\limits_{t=1}\limits^{T}\frac{ S^{l-1}[t]}{T} - \theta^{l}\sum\limits_{t=1}\limits^{T} \frac{S^{l}[t]}{T}
\end{align}
where $\sum\nolimits_{n=1}^{T} S^{l}[t]/T$ stand for the spiking rates $r^{l}$.
Thus, we can reformulate the Eq. \ref{Eq:sr_eq} as: 
\begin{align}
        \label{Eq:sr4ll} r^{l} = \frac{\mathbf{W}^{l}r^{l-1}}{\theta^{l}} - \epsilon^{l}
\end{align}
where $\epsilon^{l}$ equals $(V^l[T]-V^l[0])/(\theta^{l} \cdot T)$. 
Given equivalent input to both ANN and SNN at layer $l$, the output can be expressed as $x^l=\mathbf{W}^{l}r^{l-1}/\theta^{l}=\mathbf{W}^{l}a^{l-1}$.
The absolute conversion error, which pertains to the comparison between the converted SNN and ANN at layer $l$, can be expressed as:
\begin{align}
        \label{Eq:abs_err} {Err}^{l} = r^{l}-a^{l} = x^l - \epsilon^{l} - h(x^l)
\end{align}
The conversion error ${Err}^{l}$ is nonzero when $V^l[T]-V^l[0]\neq0$ and $x^l>0$.
Firstly, we can employ a floor function to discretize the continuous activation values $a^l$ into spiking rate representations $r^l$.
Secondly, since ReLU-generated activations can be $a^{l}>1$, whereas the spiking rate $r^{l}$ is constrained within $\{0/T,1/T,\cdots,T/T\}$, this discrepancy will result in clipping error as illustrated in Fig. \ref{fig:quant_err}.
Here, $a^{l}$ can be remapped to $r^{l}$ by floor function and spike rate clipping technique, which is named as the quant-clip function:
\begin{align}
        \label{Eq:clip_act} r^{l} = clip\Big( \frac{1}{T} \Big\lfloor\frac{a^lT}{\lambda^l}\Big\rfloor, 0, 1 \Big)
\end{align}
where $ \lfloor\cdot \rfloor$ denotes the floor function, $\lambda^l$ represents the maximum activation value ($max(a^l)$).
Therefore, we can adopt Eq. \ref{Eq:clip_act} as the activation function for ANN to compensate for conversion error. 
However, there are still numerous errors that cannot be ignored.

It can be found the conversion error is mainly caused by $\epsilon^{l}$.
Here, we assume that the residual membrane potential falls within a level insufficient to trigger spike generation, as follows: 
\begin{align}
        \label{Eq:m_range}  0\leq V^l[T]-V^l[0] < \theta^{l}
\end{align}
Under such assumption, the value of $\epsilon^{l}$ is restricted to the range $[0, 1/T)$, causing it to degenerate into the quantization error.
However, the neuronal dynamics of IF neurons can lead to non-uniform spike distributions, resulting in residual membrane potentials that violate Eq. \ref{Eq:m_range}. 
To demonstrate this phenomenon, we present three representative cases in Fig. \ref{fig:smm_and_rl}(b$\sim$d), where identical input firing rates produce different spike distribution patterns.

We suppose that the three spiking neurons in layer $l-1$ with $r^{l-1}=[0.6,0.4,0.4]$ in time-steps $T=5$
are connected to one spiking neuron in layer $l$ with $\textbf{W}^l=[1,0.5,-1]$ and firing threshold $\theta^l=1$.
Meanwhile, we also input this into the ANN to obtain the accurate output.
According to Eq. \ref{Eq:activition_base}, we can get $a^l=h([1,0.5,-1]\cdot[0.6,0.4,0.4]^T)=0.4$.
As presented in Fig. \ref{fig:smm_and_rl}(b), when the input spikes from the previous layer $l-1$ are relatively uniform, the firing rate is exactly $r^l=2/5=0.4$ which equals $a^l$, and the residual membrane potential is precisely $0$. 
However, as evidenced by Fig. \ref{fig:smm_and_rl}(c$\sim$d), when the spike firing of layer $l-1$ is irregular, there will be an error between the output spike rate and $a^l$.
Specifically, Fig. \ref{fig:smm_and_rl}(c) shows a residual potential greater than 0, indicating incomplete spike firing which results in fewer spikes being emitted. 
Conversely, Fig. \ref{fig:smm_and_rl}(d) depicts spikes being emitted prematurely and excessively, leading to a negative residual potential.
Both scenarios reflect that errors in the firing rate will occur when the assumption condition Eq. \ref{Eq:m_range} for the residual membrane potential does not hold.

The quantization error has unavoidable impact on various vision tasks. 
As shown in Fig. \ref{fig:smm_and_rl}, the ReLU function of ANN can output continuous values, whereas the quant-clip function merely maps values in the range of $[0,\theta^l]$ onto discrete spike rates.
However, for regression tasks, we believe that lower time-steps may fail to accurately represent some regression outputs. 
Here, we choose YOLO's anchor box mechanism \cite{redmon2016you} as a case study for dimensional regression, where width and height predictions are computed through the following formula:
\begin{align}
        \label{Eq:yolo_lx}  
        w,h &=exp(W^lr^{l-1})\times anchor(w,h)
\end{align}
We assume that the predefined $anchor(w,h)=[1,1]$ (the blue box in Fig. \ref{fig:quant_err}), and the output layer has two neurons with $\textbf{W}^l=[1, -1]$.
Under different quantized time-steps, the range that the final detection box can cover is illustrated in Fig. \ref{fig:quant_err} with red color.
Since the sizes of bounding boxes for detection tasks vary widely, at low time-steps, it inevitably leads to a mismatch between the prediction box and the ground truth, thereby reducing detection accuracy.
For instance, at low time-steps $T=4,8$ in Fig. \ref{fig:quant_err}, the sizes of prediction boxes cannot accurately match our assumed ground truth (Green box in Fig. \ref{fig:quant_err}) perfectly.
Meanwhile, in the experimental section \ref{sec:exp}, we also present data demonstrating the accuracy loss caused by quantization error at low quantized time-steps.

\section{Methods}\label{sec:method}
Based on the above analysis, we propose tdIF Neuron toward lossless conversion with low time-steps.
Overall, in this chapter, we initiate our approach with a source neural network that employs the trainable quant-clip function as its activation function, which enables the transition of this network to its corresponding SNN version.
The primary cause of the error in the SNN after applying the quant-clip function conversion is due to the residual potential and quantization error.
Basically, we can reduce these errors by increasing the time-steps $T$ of quant-clip function.
However, following the conversion of the network to an SNN, increasing $T$ not only diminishes the inference efficiency of SNN but also elevates power consumption.
Therefore, secondly, we introduce the delay-spike rules in tdIF to reduce errors caused by irregular spiking and propose an inference pipeline to minimize the operational delays associated with delay-spike.
Finally, we will present how our tdIF utilizes information from time-step to significantly reduce the time-steps in SNNs while also minimizing quantization errors.

\subsection{ANN-SNN Conversion with Quant-Clip Function}

\begin{figure*}[htbp]
	\begin{center}
		\includegraphics[width=0.9\linewidth]{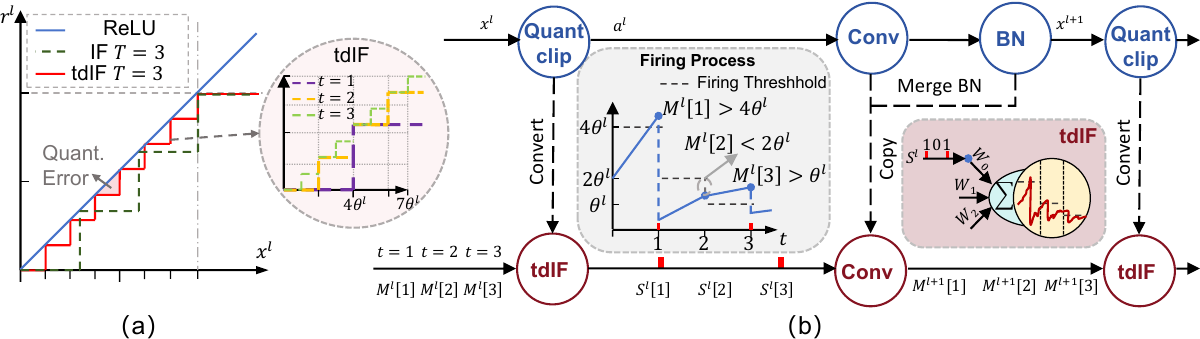}
	\end{center}
	\caption{(a) Comparison between the firing rates of the IF neuron and the tdIF at the same time-steps setting ($T=3$). By using time-steps to represent the positional information of binary numbers, our 3 time-steps tdIF achieves stronger information expressive capability equivalent to that of IF neurons within 7 time-steps. 
    (b) Overall conversion framework and the dynamics of tdIF neuron.}
	\label{fig:tdIF}
\end{figure*}

The foundational premise of ANN-SNN conversion establishes a rate-coding equivalence between spiking neural dynamics and analog activation patterns. 
To minimize the conversion error, we integrate quantization and clipping constraints into ANN training by employing modified activation functions (quant-clip).
To distinguish it from the parameters in SNNs, the activation function $\overline{h}(\cdot)$ used in ANN training can be reformulated based on the quant-clip function Eq. \ref{Eq:clip_act} as follows:
\begin{align}
        \label{Eq:clip_act_ann} \hat{a}^{l} &=\overline{h}(x^l) =\lambda^l a^{l} =\lambda^lclip\Big( \frac{1}{L} \Big\lfloor\frac{x^lL}{\lambda^l}\Big\rfloor, 0, 1 \Big)
\end{align}
where the hyperparameter $L$ denotes quantized time-steps of ANNs. 
The maximum value $\lambda^l$ of $a^l$ in ANNs is designated as a trainable parameter, 
aiming to achieve a closer mapping relationship with the firing rate $r^l$  in SNNs.

With this new activation function, we can prove that the estimated conversion error between SNNs and ANNs is zero.
When an ANN with activation function Eq. \ref{Eq:clip_act_ann} is converted to an SNN with the same weight.
According to Eq. \ref{Eq:abs_err}, the conversion error can be written as:
\begin{align}
    \label{Eq:cvt2snn_err} 
    \begin{split}
         {Err}^l &= r^l-a^l  \\
         &= clip\Big( \frac{1}{T} \Big\lfloor\frac{x^lT}{\theta^l}\Big\rfloor, 0, 1 \Big) - clip\Big( \frac{1}{L} \Big\lfloor\frac{x^lL}{\lambda^l}\Big\rfloor, 0, 1 \Big)
    \end{split}
\end{align}
If we have the conditions $T = L, \, \theta^l = \lambda^l$, then the conversion error ${Err}^l =  0 $.

As shown in Fig. \ref{fig:tdIF}(b), when completing the source ANN training with quant-clip, we integrated Conv (convolution) and BN (batch-normalization) to eliminate unnecessary multiplication operations. 
Taking a single neuron as an example, we introduce the merging scheme of Conv and BN layers. 
Given an input of last layer $\hat{a}^l$, as the Conv layer precedes BN, a common architecture in ANNs, the output after Conv transformations can be expressed as:
\begin{align}
        \label{Eq:conv_bn} 
        y &= \mathbf{W}^{l}\hat{a}^{l},    x^{l+1} = \gamma^l \frac{y-\mu}{\sqrt{\sigma^2+\epsilon}} +\beta^l
\end{align}
where $\mu$ and $\sigma^2$ denote the mean and variance of data, respectively. $\gamma^l$ and $\beta^l$ are learnable affine parameters, and $\epsilon$ denotes a numerical stabilizer to ensure non-zero denominators. 
Meanwhile, in Eq. \ref{Eq:clip_act_ann}, parameter $\lambda^l$ is introduced for adapting to the relationship between $a^l$ and spike firing rates $r^l$.
Crucially, both  $\lambda^l$ and  $\lambda^{l-1}$ of the previous activation function can be pre-computed and fused into Conv weights during inference, thereby preserving the binary (0/1) spike transmission characteristics in the converted SNN.
All of these parameters are fixed during inference.
The membrane potential at time-step $t$ is therefore computed with $\hat{\mathbf{W}}^{l}S^{l}[t]+\hat{\textbf{B}}^l$.
The new parameters of the Conv layer can be written as follows
\begin{align}
        \label{Eq:conv_bn_merge} 
        \hat{\mathbf{W}}^{l} &=  \frac{\lambda^{l-1} \gamma^l}{\lambda^l\sqrt{\sigma^2+\epsilon}}\mathbf{W}^{l} \\
        \hat{\textbf{B}}^l &= -\frac{\mu \gamma^l}{\lambda^l\sqrt{\sigma^2+\epsilon}} +\frac{\beta^l}{\lambda^l}
\end{align}
Subsequently, this weight configuration strategy effectively eliminates redundant multiplication operations in the SNN implementation.

\subsection{Delay-Spike}

Due to the irregular firing of spikes leading to residual potential error, we design a two-stage delay-spike algorithm to achieve accurate firing rate and prevent the occurrence of residual membrane potential.
The algorithmic rule is shown in Alg. \ref{alg:delay_spiking}.
At stage $1$, we first accumulate the input value until the time $t$ reaches time-delay $T_{delay}$, thereby yielding the current accumulated membrane potential $\overline{M}$ of neurons.
Then, when time $t$ reaches $T_{delay}$, the neuron begins to emit spikes. 
In the second stage, based on the delay step length, the remaining membrane potential will be directly utilized to calculate and emit the remaining spikes.

\begin{algorithm}
    \caption{Delay-Spike}
    \label{alg:delay_spiking}
    \begin{algorithmic}[1]
    \INPUT $T_{delay}$, $\overline{M}^{l}=V^{l}[0]$, SNNs infer mode is set as multi-step mode, the spike of previous layer $\{S^{l-1}[1],S^{l-1}[2],\cdots,S^{l-1}[T]\}$
    \OUTPUT $S^{l}$
    \STATE STAGE 1: Accumulating potential and delay spiking
    \FOR{$ t = 1,\cdots T$}
        \STATE $\overline{M}^{l} =\overline{M}^{l} + \mathbf{W}^{l}S^{l-1}[t]$
        \IF{$t>T_{delay}$}
        \STATE $S^{l}[t-T_{delay}] =\Theta(\overline{\textbf{M}}^{l} -\theta^{l})$
        \STATE $\overline{M}^{l} =\overline{M}^{l}-\theta^{l}S^{l}[t-T_{delay}]$
        \ENDIF
    \ENDFOR
    \STATE STAGE 2: Firing the remaining spikes
    \FOR{$ t= T-T_{delay},\cdots T$}
        \STATE $S^{l}[t] =\Theta(\overline{\textbf{M}}^{l} -\theta^{l})$,  $\overline{M}^{l} =\overline{M}^{l}-\theta^{l}S^{l}[t]$
    \ENDFOR 
    \end{algorithmic}
\end{algorithm}

When the time-delay $T_{delay}=T$, the membrane potential $\overline{M}^{l} =\mathbf{W}^{l} \sum_{t=1}^{T}\frac{ S^{l-1}[t]}{T}$.
If we assume that the input to this layer is unbiased, according to Eq. \ref{Eq:clip_act_ann}, the current accumulated membrane potential is $\overline{M}^{l}=T\times x^l$. 
Given the conditions $\lambda^l=\theta^l$ and $0\le x^l <  \theta^l$, then spikes are generated based on the $\overline{M}^{l}$ until it diminishes to a level where further spike generation is no longer feasible, therefore, it can be asserted that hypothesis Eq. \ref{Eq:m_range} remains valid as :
\begin{align}
        \label{Eq:res_mem} 0\le \overline{M}^{l}-\theta^{l}\sum_{t=1}^{T}{S^{l}[t]}<\theta^{l}
\end{align}

\begin{figure}[t]
	\begin{center}
		\includegraphics[width=1.0\linewidth]{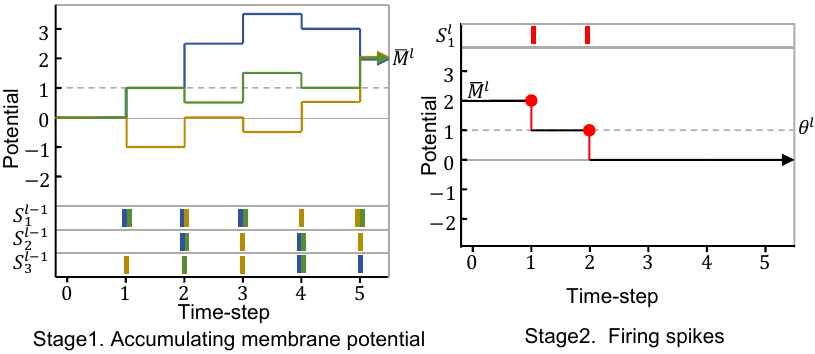}
	\end{center}
    \caption{Example of delay-spike process under $T_{delay}=5$.}
	\label{fig:delayspike}
    \vspace{-0.3cm}
\end{figure}

Under such hypothetical conditions, we recalculate the spike generation according to the firing pattern depicted in Fig. \Ref{fig:smm_and_rl}(b$\sim$d) with this delay-spike method.
The spike generation patterns for three cases are shown in  Fig. \ref{fig:delayspike}.
In this figure, during stage $1$, the cumulative membrane potential at the final time-step remains consistent across all three cases, as the spike-firing rate of the preceding layer is identical, regardless of the specific firing patterns.
Meanwhile, results in Fig. \ref{fig:delayspike} (stage $2$) confirm zero residual membrane potential after the final spiking event under our delay-spike method, which minimizes conversion errors caused by residual potential.

\subsection{Pipelining of Delay-Spike}

Similar to work \cite{liu2022spikeconverter}, adopting a time-step-driven inference approach, SNNs based on the delay-spike strategy inevitably require extended inference time-steps to execute a single inference. 
Inspired by the allocation of different processing units to different layers to minimize memory access costs, we employ inter-layer direct transmission and inter-sample pipelining to reduce the time latency of individual samples while maximizing the throughput of multiple samples.

\begin{figure}[t]
	\begin{center}
		\includegraphics[width=1.0\linewidth]{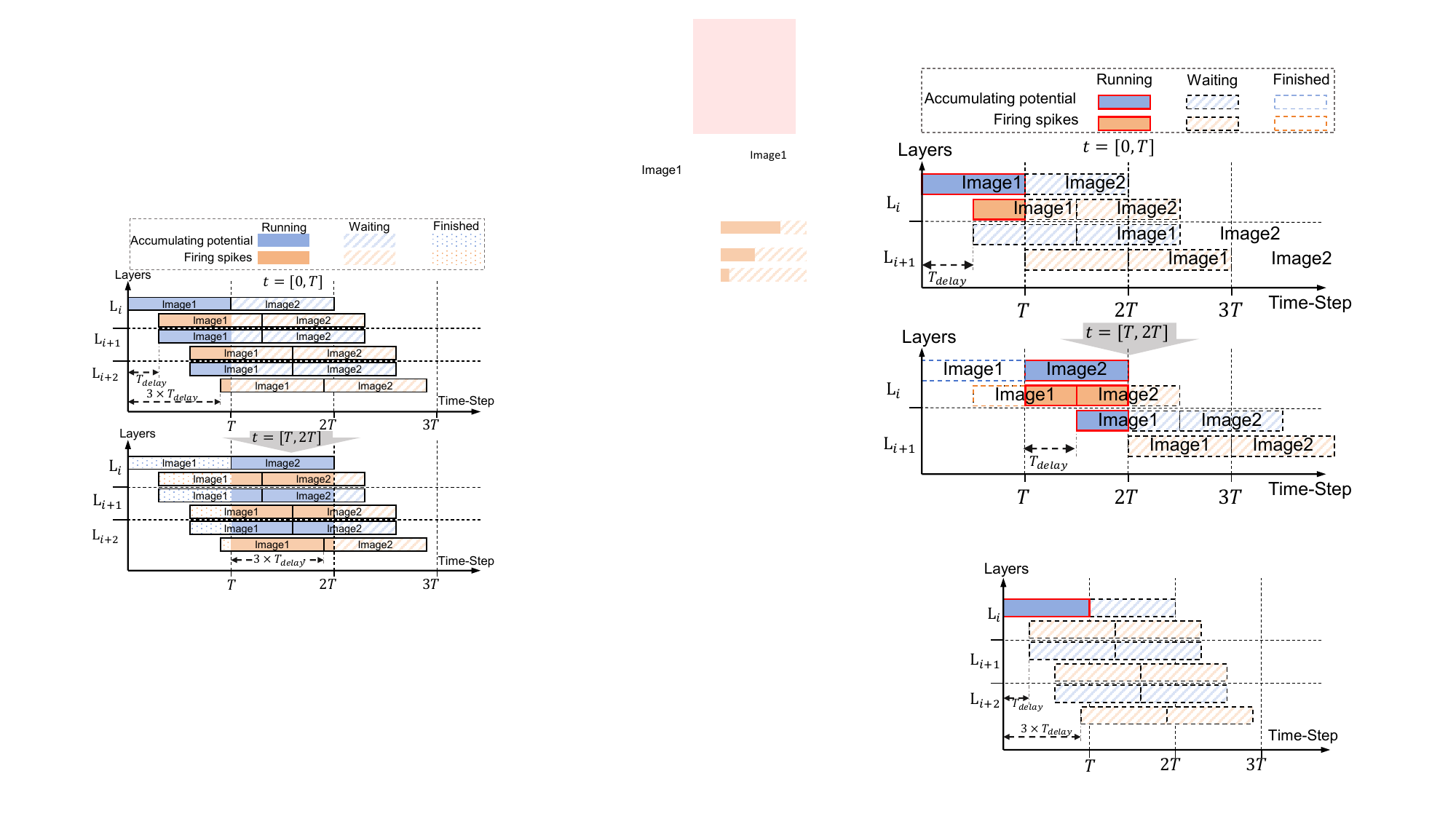}
	\end{center}
	\caption{An example of the pipline for SNN with Delay-Spike.}
	\label{fig:pipeline}
    \vspace{-0.3cm}
\end{figure}

Fig. \ref{fig:pipeline} illustrates the inference pipeline for a 3-layer SNN with time-delay. 
The horizontal axis corresponds to the temporal dimension of inference (time-steps), with the vertical axis enumerating the network layers in sequential order.
The diagram displays the distinct operational states of each layer’s processing unit during membrane potential accumulation and spike firing. 
We set the duration of both phases (accumulation and firing) for each layer to the same value $T$. 
During time-step $t=[0,T]$, layer $i$ begins spike firing synchronously at $t=T_{delay}$ after completing partial membrane potential accumulation. 
Its output is directly transmitted to layer $i+1$ for simultaneous potential accumulation.
Inter-layer direct transmission reduces the inference latency of individual samples. 
Based on this pipeline, for an SNN with $n$ layers, the total inference time-steps are $T+n\times T_{delay}$. 
Additionally, sample $2$ begins the accumulation phase in layer at time-step $T$ to fully utilize the resources of layer $i$, thereby avoiding idle hardware waste. 
Notably, for multi-time-steps layer-driven inference methods, the delay-spike strategy does not increase the computational complexity of intra-neuron operations, and the on-chip inference time remains largely unaffected.

\subsection{Temporal-Dependent IF Neuron}

For current frequency-based conversion methods, the time-steps are often unrelated to each other in terms of sequence. 
The emission of spikes only affects the magnitude of the frequency, regardless of which time-step the spike is emitted in.
To fully leverage the temporal information embedded within each time-step, we propose a neuron model with a dynamic threshold that adapts based on the time-step information.
In this neuron, we adopt a binary coding approach, where time-steps are represented as binary bit information.
Therefore, based on Eqs. \ref{Eq:charge_func}$\sim$\ref{Eq:reset_func}, the membrane potential accumulation rule and spike emission rule for neurons that utilize time-step information can be expressed as follows:

\begin{align}
        \label{Eq:charge_func_tdIF} M^{l}[t] &=V^{l}[t-1] + c[t]\mathbf{W}^{l}S^{l-1}[t]\\
        \label{Eq:fire_func_tdIF} S^{l}[t] &=\Theta(M^{l}[t]-c[t]\theta^{l})\\
        \label{Eq:reset_func_tdIF} V^{l}[t] &=M^{l}[t]-c[t]\theta^{l}S^{l}[t]
\end{align}

Where $c[t]=2^{T-t}$ represents the temporal information of spike binary coding at time-step $t$.
Eq. \ref{Eq:charge_func_tdIF} shows that the input membrane potential is multiplied by $c[t]$  according to the time-step.
The spiking process Eq. \ref{Eq:fire_func_tdIF} and the soft reset mechanism Eq. \ref{Eq:reset_func_tdIF} 
utilize dynamic firing thresholds at different time-steps to calculate the operation of spike firing and membrane potential reset.
The firing process is also illustrated in Fig. \ref{fig:tdIF}(b).
Therefore, Eq. \ref{Eq:sr_eq}  can be modified as follows: 
\begin{equation}\label{tdIF_rate}
        \begin{aligned}
            \frac{V^l[T]-V^l[0]}{T^{\prime}} &= \mathbf{W}^{l} \sum\limits_{t=1}\limits^{T}\frac{ c[t]S^{l-1}[t]}{T^{\prime}} - \theta^{l}\sum\limits_{t=1}\limits^{T} \frac{c[t]S^{l}[t]}{T^{\prime}}
        \end{aligned}
\end{equation}
where $T^{\prime}=\sum_{t=1}^{T}c[t]=2^{T}-1$.
Hence, the $r^l=\sum\nolimits_{n=1}^{T} c[t]S^{l}[t]/T^{\prime}$ of the tdIF can also be mapped to $a^l$ of the ANN.

Furthermore, according to the rules of binary encoding, our tdIF is capable of representing more information with fewer time-steps. 
A $3$ time-steps tdIF neuron example, shown in Fig. \ref{fig:tdIF}(a), demonstrates how our tdIF combines temporal information to represent binary coding.
Fig. \ref{fig:tdIF}(a) presents comparative activation curves demonstrating the input current $x^l$ to firing-rate relationships for three neuron models: conventional ReLU, standard IF, and our proposed tdIF.
As evident from this figure, our tdIF model can match information equivalent to approximately $7$ time-steps within just $3$ time-steps, effectively reducing the quantization error compared to ReLU.
Meanwhile, within the same $3$ time-steps, compared to IF neurons, our tdIF demonstrates superior expressive capability, enabling it to maintain high precision while ensuring low-latency characteristics.

\subsection{Decoding of tdIF.}\label{meh:od}

In regression tasks, the final layer of conventional ANNs typically generates numerical values through direct convolutional or fully-connected layers.
When converted to SNNs, the final output corresponds to the membrane potential value defined as $V^o[t]=\textbf{W}^oS^l[t]$, where $t \in [1,T]$ , $\textbf{W}^o$ is the weights of output layer and $S^l[t]$ is the input spikes.
In implementations using  IF neuron models, the equivalent output value relative to the source ANN is expressed becomes:
\begin{align}
        \label{Eq:decode4IF} 
        O_{IF} &= \frac{ \sum_{t=1}^{T}{V^o[t]} }{T}
\end{align}

When adopting the tdIF model, the temporal information embedded in discrete time-steps is encoded as positional indices in binary representations. 
Consequently, the final output membrane potential exhibits varying contributions from different time-step. 
The decoding methodology can be mathematically formulated as follows:
\begin{align}
        \label{Eq:decode4tdIF} 
        O_{tdIF} &= \frac{ \sum_{t=1}^{T}{V^o[t] \times 2^{T-t}} }{2^T-1}
\end{align}

where $O_{tdIF}$ denotes the decoded output, $2^{T-t}$ represents time-dependent weighting coefficients, and corresponds to the membrane potential value at time-step. 
Thus, the final output obtained through this method is equivalent to the original output of source ANN, and it can be applied as the final decoding component for various types of regression tasks.

\section{Experiments}\label{sec:exp}
To demonstrate the effectiveness and efficiency of our algorithm for visual detection tasks, we compare our tdIF method with existing state-of-the-art approaches for image object detection and lane line detection tasks in this section.

We evaluate object detection performance across two benchmark datasets: PASCAL VOC \cite{everingham2010pascal} and MS COCO 2017 \cite{wang2023toward}. 
The baseline architecture adopts a streamlined YOLOv3-Tiny implementation \cite{farhadi2018yolov3}, modified for SNN compatibility through two key adaptations: substitution of leakyReLU activations with standard ReLU, and replacement of max pooling operations with average pooling.
To assess scalability, we additionally implement a YOLOv3 variant with a backbone of ResNet34 \cite{he2016deep}.
We compare our method with the contemporary ANN-SNN conversion methods \cite{kim2020spiking,kim2020towards} and the Two-Stage Optimization method \cite{wang2023toward}, Fast-SNN \cite{hu2023fast}.

For lane line detection task, we also conducte experiments on two datasets: Tusimple \cite{pizzati2020lane} and CULane \cite{pan2018spatial}. 
We mirror the head structure from CondlaneNet  \cite{liu2021condlanenet} as our lane detection head.
In Condlane-head, we adopt the membrane potential of the final output layer to represent the parameter map, which serves as the parameter for dynamic convolutional kernels.
Meanwhile, the spike-rate is utilized to characterize the location map and offset map.
Furthermore, our experiment incorporates two distinct ResNet configurations - ResNet18 and ResNet34 - as feature extraction networks, systematically assessing the depth adaptability of our conversion methodology in lane detection scenarios.

More information regarding the experimental setting can be found in Appendix \ref{apx:es}. 

\begin{figure}[t]
	\begin{center}
		\includegraphics[width=1.0\linewidth]{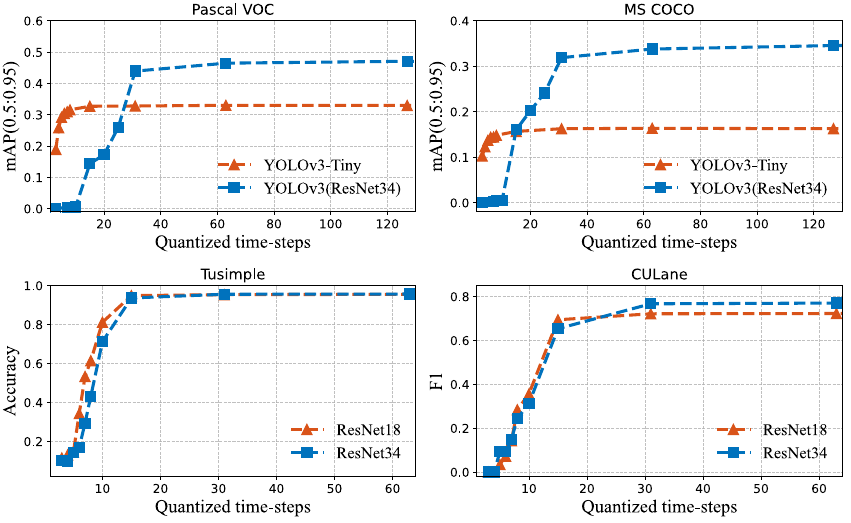}
	\end{center}
	\caption{Accuracy gap between source ANN with Quant-Clip Function at different quant time-steps.}
	\label{fig:acc2ts}
    \vspace{-0.4cm}
\end{figure}

Firstly, we investigate the accuracy loss due to quantization error at different time-steps. 
As shown in Fig. \ref{fig:acc2ts}, we employ various quantization time-steps on four identical datasets. 
For all tasks across the datasets, when the time-steps is less than $8$, the task accuracy is significantly lower than that of ANNs. 
When the time-step exceeds 32, the accuracy basically converges and approaches the values of ANNs.
Fig. \ref{fig:acc2ts} also indicates that deep networks require relatively more time-steps to achieve accuracy convergence. 
This figure also demonstrates the accuracy loss due to quantization at low time-steps, as mentioned in 
sec. \ref{sec:cvt_err}.

\subsection{Performance}
In this section, all results of our method employ the configuration of $T_{delay} = T$.
\subsubsection{Object Detection}

\begin{table*}[t]
        \caption{Performance comparison for object detection task on PASCAL VOC and MS COCO.
        }
        \centering
        \tabcolsep=0.1cm     
        
        \begin{tabular}{l|l|c|c|c|c|c|c}
        \hline
        \hline
       \multirow{2}{*}{Method}   & \multirow{2}{*}{Architecture}  & \multirow{2}{*}{\makecell{Param\\(M)}}   & \multirow{2}{*}{Time-steps}  &  \multicolumn{2}{c|}{PASCAL VOC} &  \multicolumn{2}{c}{MS COCO}  \\  \cline{5-8}
          &  &   &  & AP@0.50:0.95 & AP@0.50    & AP@0.50:0.95 & AP@0.50 \\\hline\hline
        
         \multirow{2}{*}{\makecell{Spiking-YOLO \\ \cite{kim2020spiking}} }  & YOLOv3-Tiny       & 8.85    & 8     & 0.00  & 0.23  & 0.00   & 0.11\\
                                & YOLOv3-Tiny       & 8.85    & 2000  & 26.9  & 59.9  & 12.51  & 25.78\\\hline
        \multirow{2}{*}{\makecell{Spike Calibration\\\cite{li2022spike}}}& YOLO(ResNet50)  & 45.24   & 64    & 31.75  & 63.85 & 16.17     & 33.12 \\
                                & YOLO(ResNet50)  & 45.24    & 512   & 44.36     & 75.21 & 23.50     & 45.42 \\\hline
        
        \multirow{2}{*}{\makecell{Two-Stage Optimization\\\cite{wang2023toward}}} & YOLOV3-Tiny       & 8.85    & 20    & 28.5  & 64.1  & -      & - \\
                                & YOLOV3-Tiny       & 8.85    & 40    & 30.7  & 66.4  & -      & - \\\hline
        \multirow{4}{*}{\makecell{Fast-SNN\\\cite{hu2023fast}}} & Tiny YOLO         & 15.86    & 7     & -     & 52.83 & -      & 26.49\\
                                & Tiny YOLO         & 15.86    & 15    & -     & 53.17 & -      & 27.59\\
                                & YOLOv2(ResNet34)  & 55.01    & 7     & -     & 73.43 & -      & 41.89\\
                                & YOLOv2(ResNet34)  & 55.01 & 15    & -     & 76.05 & -      & 46.40\\ \hline

        \multirow{6}{*}{Ours}   & YOLOv3-Tiny       & 8.85   & 5     & 32.33 & 63.96                    & 16.26  & 36.12\\
                                & YOLOv3-Tiny       & 8.85   & 7     & 32.76 & 64.43                    & 16.26  & 36.25\\
                                & YOLOv3-Tiny       & 8.85   & 8     & 32.80 & 64.53                    & 16.30  & 36.29\\
                                & YOLOv3(ResNet34)  & 41.98  & 5     & 43.46 & 74.41                    & 31.93  & 55.73\\
                                & YOLOv3(ResNet34)  & 41.98  & 7     & \textbf{47.33} & \textbf{78.80}  & 34.50  & 58.89\\
                                & YOLOv3(ResNet34)  & 41.98  & 8     & 46.46 & 78.46            & \textbf{35.16} & \textbf{59.78}\\\hline\hline

        \end{tabular}
        \label{table:bbox_performance}
        \vspace{-0.4cm}
\end{table*}

\begin{figure*}[htpb]
	\begin{center}
		\includegraphics[width=0.95\linewidth]{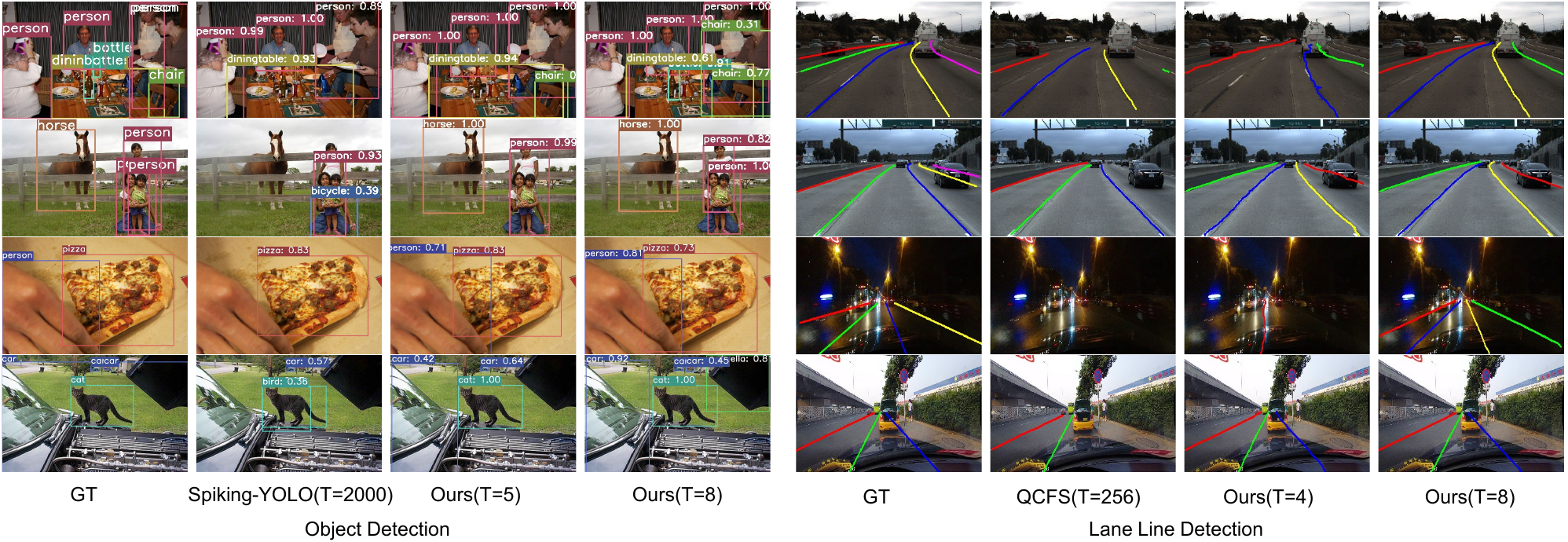}
	\end{center}
	\caption{Visual quality comparison of object detection results with the network architecture YOLOv3(ResNet34) and lane line detection with ResNet34+Condlane. 
    }
	\label{fig:VQC}
    \vspace{-0.4cm}
\end{figure*}

We summarize and compare the performance in Table \ref{table:bbox_performance}.
We include the Spiking-YOLO \cite{kim2020spiking}, spike Calibration \cite{li2022spike}, Two-Stage Optimization \cite{wang2023toward} and Fast-SNN \cite{hu2023fast} for comparison. 
These numbers are either taken from the corresponding papers or recalculated using their open-source code.
Our algorithm achieves optimal detection accuracy in both datasets. 
Even with around $5$ time-steps and similar model parameters, our method significantly outperforms others. 
Spiking-YOLO barely has accuracy at short time-steps, and even when other time-steps exceed 2000, it remains inferior to our YOLOv3-tiny model ($5.4$ percentage points lower on Pascal VOC and $3.35$ percentage points lower on COCO under mean $AP@0.50:0.95$). 
Spike Calibration, despite having a notably superior feature extractor, still requires up to $512$ time-steps to match the accuracy of our model using ResNet34. 
The Two-Stage Optimization method also requires a higher number of time-steps, exceeding $40$, to outperform us on VOC but is approximately $1.13$ percentage points lower than our tiny model on  mean $AP@0.50:0.95$. 
Fast-SNN has lower time-stepp requirements, but its method still falls below our algorithm. 
At the same time-steps of $7$, our algorithm significantly outperforms Fast-SNN on both datasets.
Additionally, Fast-SNN is a two-stage model that demands more computational resources.
Meanwhile, our model experiences lower accuracy loss with shallower layers at different steps(with YOLOV3-Tiny, $0.47$ percentage loss on VOC, $0.04$ percentage loss on COCO), but higher loss with deeper layers (with YOLOV3-Resnet34, $3.0$ on VOC, $3.23$ on COCO).

\begin{table*}[htbp]
        \caption{Performance comparison for Lane line detection task on Tusimple and CULane.
        }
        \centering
        \tabcolsep=0.1cm     
        \begin{tabular}{l|c|c|p{1.7cm}<{\centering}|p{1.7cm}<{\centering}|p{1.7cm}<{\centering}|p{1.7cm}<{\centering}|p{1.7cm}<{\centering}|p{1.7cm}<{\centering}}
        \hline\hline
       \multirow{2}{*}{Method}   & \multirow{2}{*}{Backbone}  &  \multirow{2}{*}{Time-steps}  &  \multicolumn{3}{c|}{Tusimple} &  \multicolumn{3}{c}{CULane}  \\ \cline{4-9}
          &  &   & Acc(\%) $\uparrow$ & FP(\%) $\downarrow$ & FN(\%)  $\downarrow$  & F1(\%) $\uparrow$ & Prec.(\%) $\uparrow$& Recall(\%) $\uparrow$ \\\hline\hline 

        \multirow{4}{*}{\makecell{QCFS\\\cite{bu2023optimal}}} & \multirow{2}{*}{ResNet18}& 128 & 60.18  & 35.96 & 65.24 & 37.9 & 93.71  & 23.79\\
                                 &   & 256  & 92.25    & 2.25   & 9.03  & 61.33     & 90.95 &46.26\\ \cline{2-9}
        & \multirow{2}{*}{ResNet34}  & 128  & 64.01    & 21.87  & 56.89 & 7.79      & 83.83 &4.09\\
                                 &   & 256  & 93.88    & 3.29   & 6.82  & 48.35     & \textbf{94.17} &32.53\\ \hline

        \multirow{8}{*}{Ours}& \multirow{4}{*}{ResNet18}    & 4    & 94.80 & 3.59 & 5.29    & 69.32 & 80.67  & 60.77\\
                                                        &   & 5    & 95.35 & 2.53 & 4.27    & 72.17 &80.30  &65.53\\
                                                        &   & 7    & 95.37 & 2.41 & 4.31    & 72.23 &79.82  &65.97\\
                                                        &   & 8    & 95.29 & 2.63 & 4.36    & 72.35 & 81.15 &65.27 \\ \cline{2-9}
                               & \multirow{4}{*}{ResNet34}  & 4    & 93.55 & 6.02 & 7.89    & 65.38 &78.80 &55.87\\
                                                        &   & 5    & 95.48 & 2.53 & 4.14    & 76.65 &81.29 &72.52 \\
                                                        &   & 7    & 95.53 &\textbf{2.33}&\textbf{3.98}    & 77.17 &81.93 &72.93 \\
                                                        &   & 8    &\textbf{95.56}& 2.47 & 4.13 &\textbf{77.61}&82.20&\textbf{73.51}\\\hline\hline

        \end{tabular}
        \label{table:lane_performance}
        \vspace{-0.3cm}
\end{table*}

\subsubsection{Lane Line Detection}

We summarize and compare the performance in Table \ref{table:lane_performance}.
Since we are the first to adopt the ANN-SNN conversion algorithm for lane line detection task, 
we apply the QCFS method to lane detection tasks for comparison. 
Both methods use CondLane as the detection head and refer to CondLane's training strategies for weight optimization. 
Our algorithm achieves high detection accuracy within a short number of time-steps, reaching over $94\%$ on the TuSimple dataset with 4 time-steps and over $69\%$ on the CULane dataset. 
Experimental results demonstrate that the model achieves consistent accuracy when the time-steps extends beyond $5$.
However, due to poor handling of cumulative errors, OCFS has almost poor performance when the number of time-steps is less than $128$ and requires a larger number of time-steps (greater than $256$) to achieve corresponding detection accuracy. 
Given the limited number of samples in the TuSimple dataset (around 3000 in total), using ResNet18 and ResNet34 yields similar performance, with a difference in detection accuracy of only $0.27\%$ at $8$ time-steps. 
In contrast, the difference in performance between these two networks on the CULane dataset is around $5.26\%$. 
Additionally, deeper networks experience greater losses at low latency, with ResNet34 showing a difference in accuracy of $2.01\%$ between $8$ and $4$ time-steps, while ResNet18 only shows a difference of $0.49\%$ on the TuSimple dataset.

In Fig. \ref{fig:VQC}, we further provide visual results based on ResNet34, encompassing both object detection and lane detection. 
As illustrated, our SNN is capable of producing detection results close to the ground truth with lower time-steps. 
For instance, our algorithm with a latency of $8$ can differentiate some overlapping objects (e.g., bottle and person) and precise lane lines with higher accuracy.
Appendix \ref{apx:addres} contains more information on the detection accuracy of different quantized time-steps with time-steps $[3,4,5,7,8,10]$ across four datasets.

\subsection{Ablation Studies}

\begin{figure}[htbp]
	\begin{center}
		\includegraphics[width=0.85\linewidth]{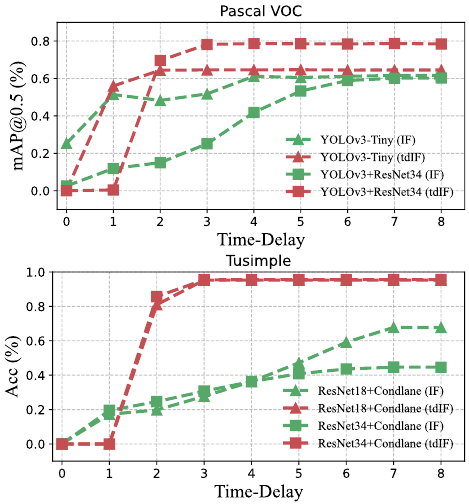}
	\end{center}
	\caption{Ablation study with/without tdIF and different time-delay.}
	\label{fig:ablation}
        \vspace{-0.3cm}
\end{figure}

\begin{table*}[t]
        \caption{
        Component of IF neuron and tdIF neuron on digital neuromorphic processor.
        }
        \centering
        \tabcolsep=0.3cm     
        \renewcommand{\arraystretch}{1.0}
        \begin{tabular}{l|l|l}
        \hline
        IF  neuron                & tdIF     neuron                  & Description \\ \hline
        \multicolumn{2}{c|}{Component for Accumulating potential.}&\\ \hline
        \textcolor{black}{MLD}(R0, ADD1, 1)&\textcolor{black}{MLD}(R0, ADD1, 1)&load weight $W^l$\\
        \textcolor{black}{MLD}(R1, ADD2, 0)&\textcolor{black}{MLD}(R1, ADD2, 0)&load state $V^l[t-1]$\\
         &\textcolor{black}{SHL}(R0, R0, R2)&$c[t]\times w_{ij}$ via bitwise shifting, R2=$T-t$ \\
        \textcolor{black}{ADD}(R1, R0, R1)& \textcolor{black}{ADD}(R1, R0, R1)& $M^l[t] = V^l[t-1] + W^l$\\
         \textcolor{black}{MST}(ADD2, R1, 1)&\textcolor{black}{MST}(ADD2, R1, 1)&store R1 in $M^l[t]$\\ \hline
        12.7 &13.9 & Energy for Accumulating potential (pJ)\\ \hline
        \multicolumn{2}{c|}{Component for Firing spikes.}\\ \hline
        \textcolor{black}{MLD}(R0, ADD2, 0)& \textcolor{black}{MLD} (R0, ADD2, 0)&load state $M^l[t]$\\
        \textcolor{black}{GTH}(R2, R0, R1)&\textcolor{black}{GTH}(R2, R0, R1)&generate spike, IF:$\Theta(M^l[t] - \theta^l)$, tdIF:$\Theta(M^l[t] - \hat{\theta}^l[t])$\\
        \textcolor{black}{MUL}(R3, R2, R1)&\textcolor{black}{MUL}(R3, R2, R1)&IF:$\theta \times S^l[t] $, tdIF:$\hat{\theta}[t] \times S^l[t]$\\
        \textcolor{black}{SUB}(R0, R0, R3)&\textcolor{black}{SUB}(R0, R0, R3)&reset state if spike\\
        \textcolor{black}{MST}(ADD2, R0, 1)&\textcolor{black}{MST}(ADD2, R0, 1)&store R0 in $V^l[t]$\\
        \textcolor{black}{EVC}(R2)       &\textcolor{black}{EVC}(R2)&capture event\\ \hline
        13.2 &13.2 & Energy for Firing spikes (pJ)\\ \hline
        
        \end{tabular}
        \label{table:ec4n}
        \vspace{-0.8cm}
\end{table*}

\begin{table}[htbp]
        \caption{\textbf{Neuron Processing Energy Consumption.} 
        }
        \centering
        \tabcolsep=0.2cm     
        \renewcommand{\arraystretch}{1.0} 
        \begin{tabular}{l|c|c}
        \hline
        Instruction         & Description                       & Energy (pJ) \\ \hline
        ADD/SUB             & \multirow{1}{*}{Arithmetic ops.}  & 1.4 \\ \hline
        GTH/MAX/MIN         & \multirow{1}{*}{Compare ops.}     & 1.2 \\  \hline
        SHL/SHR             &    Bit-wise ops.                  & 1.2 \\ \hline
        \multirow{2}{*}{EVC}& Event Capture                     & 0.5 \\ 
                            & + if generates event              &+ 1.1\\ \hline
        MLD                 & \multirow{2}{*}{Data Mem Load/Store}  & 3.7 \\ 
        MST                 &                                   & 3.9\\ \hline
        \end{tabular}
        \label{table:NPEC}
        \vspace{-0.8cm}
\end{table}

We conduct ablation experiments combining different neuron types (IF and tdIF) and different time-delay on Pascal VOC dataset and Tusimple dataset at an $8$ time-steps setup. 
Fig. \ref{fig:ablation} presents the results of our experiments. 
The x-axis represents different time-delay ranges from $0$ to $8$.
It can be seen that the combination of tdIF and delay-spike can effectively reduce the time-steps while achieving a higher detection accuracy. 
Meanwhile, equipped with tdIF relies more heavily on the time-delay term (our tdIF has no accuracy when $T_{delay}=0,1$).
We attribute this to tdIF’s explicit incorporation of temporal information (adjust membrane potential accumulation based on time-steps). 
In Eq. \ref{Eq:charge_func_tdIF}, the first time-step exerts the strongest influence on membrane potential due to $c[1] = 2^{T-1}$. Consequently, spikes generated with temporal deviation (either delayed or mistimed) significantly disrupt the membrane potential accumulation. 
This temporal sensitivity makes our method more dependent on precise delay-spike coordination compared to conventional frequency-based IF strategies. 

However, when employing deep network architectures with IF neurons, such as ResNet, they also fail to exhibit performance at  $T_{delay}=0$.
As the amplified  accumulation characteristic of tdIF enables faster membrane potential buildup during initial time-steps, our tdIF demonstrates significantly superior performance compared to models that utilize IF Neuron at $T_{delay}\ge2$ and basically achieves peak performance at $T_{delay} = 3$, whereas IF requires substantially more delayed time to reach its highest levels.
Without the adoption of tdIF, all models except for the shallow one like YOLOv3-Tiny exhibit significant losses compared to those with tdIF.
Specifically, when $T_{delay}$ is set to $8$, there is a precision loss exceeding $10$ percent.

\subsection{Computation Cost}

\subsubsection{Energy Consumption for tdIF}

\begin{table*}[t]
        \caption{\textbf{Inference Energy Consumption} 
        }
        \centering
        \tabcolsep=0.2cm     
        \renewcommand{\arraystretch}{0.9}
        \begin{tabular}{l|c|c|c|c|c|l}
        \hline
        Architecture & Time-steps & Neuron  & Performance & \makecell{Average \\Spiking rate}  & \makecell{Power\\(mJ)} &  $E/E_{ref}$ \\ \hline \hline
        \multicolumn{7}{c}{Results on Pascal VOC} \\ \hline
        \multirow{3}{*}{YOLOv3-Tiny}          &8  & IF  & 61.60  & 0.079 & 12.1      & 1.0$\times$ (reference)   \\  
                                              &32 & IF  & 64.07  & 0.078 & 47.4      & 3.9$\times$   \\      
                                              &8  & tdIF& 64.53  & 0.173 & 31.0      & 2.6$\times$   \\  \hline
        \multirow{2}{*}{\makecell{YOLOv3 \\(ResNet34)}}    &8  & IF  & 60.20  & 0.056 & 123.2     & 1.0$\times$ (reference)   \\ 
                                              &32 & IF  & 74.40  & 0.056 & 475.6     & 3.8$\times$   \\ 
                                              &8  & tdIF& 78.46  & 0.169 & 423.8     & 3.4$\times$   \\ \hline
        \multicolumn{7}{c}{Results on Tusimple} \\ \hline
        \multirow{2}{*}{\makecell{ResNet18 \\+Condlane}}   &8  & IF  & 67.75  & 0.060 & 113.4     & 1.0$\times$ (reference)   \\ 
                                              &16 & IF  & 94.87  & 0.052 & 220.8     & 1.9$\times$   \\
                                              &8  & tdIF& 95.27  & 0.165 & 271.3     & 2.4$\times$   \\ \hline
        \multirow{2}{*}{\makecell{ResNet34 \\+Condlane}}   &8  & IF  & 44.65  & 0.041 & 118.0     & 1.0$\times$ (reference)   \\ 
                                              &16 & IF  & 93.71  & 0.042 & 265.8     & 2.3$\times$   \\
                                              &8  & tdIF& 95.56  & 0.152 & 387.5     & 3.3$\times$   \\ \hline
        \end{tabular}
        \label{table:EngCom}
        \vspace{-0.4cm}
\end{table*}

Since our tdIF neuron introduces a multiplicative operation on the input membrane potential, it may incur additional energy consumption. 
Here, we first analyze the power consumption calculation of both IF and tdIF neurons.
As mentioned in \cite{tang2023open}, we have the energy consumption of the digital neuromorphic processor for each operation in Table \ref{table:NPEC}. 
Therefore, based on the operation counts of the digital neuromorphic processor, we can divide the membrane potential accumulation and spike generation of the IF neuron into two parts as shown in Table \ref{table:ec4n}. 
Due to the binary (0,1) nature of spikes, the membrane potential accumulation process is spike-driven, while spike generation is computed at every time-step. 
This yields energy consumption values of $12.7 pJ$ and $13.2 pJ$ for these two processes in a single IF neuron, respectively.

Although the tdIF neuron incorporates a multiplication operation in its membrane potential accumulation (Eq. \ref{Eq:charge_func_tdIF}), it maintains the sparse (0,1) transmission characteristics between layers and preserves the event-driven computation property. 
Consequently, only addition operations are required when computing input membrane potentials in convolutional or fully-connected layers.
For the amplification coefficient $c[t]=2^{T-t}$ in Eq. \ref{Eq:charge_func_tdIF}, we can implement it using bit-shift operations. 
Thus, during the potential accumulation process shown in Table \ref{table:ec4n}, tdIF requires only one additional left-shift operation compared to IF neuron. 
Furthermore, for the spike generation computation, we can store the threshold values per time-step as 
$\hat{\theta}^l=[2^{T-1}\theta^l,2^{T-2}\theta^l,\cdots,\theta^l]$. 
This implementation results in energy consumption values of $13.9 pJ$ and $13.2 pJ$ for the two processes in tdIF neuron, representing only a $1.2 pJ$ overall increase compared to the standard IF neuron.

\subsubsection{Inference Energy Cost}

We compared the energy consumption during model inference between our approach and models using IF neurons across two detection tasks. 
As shown in Table \ref{table:EngCom}, all models in this experiments adopted the $T_{delay}=T$ setting with $8$ time-steps. 
To ensure fair comparison under similar performance levels, we configured the IF neuron models with $32$ time-steps for object detection and $16$ time-steps for lane detection task.

The models employing tdIF neurons consistently demonstrated an average firing rate of approximately $0.15$ across all four architectures, significantly higher than the $0.06$ firing rate of conventional IF neurons. 
Consequently, under identical time-step settings, tdIF-based models consume about $2$ to $3$ times more energy than their IF counterparts.
However, to achieve comparable performance to our tdIF models, IF-based models require increased time-steps, inevitably leading to higher power consumption. 
For instance, the YOLOv3 model on Pascal VOC dataset must extend to $32$ time-steps, resulting in substantially higher power consumption than our approach. 
On the Tusimple dataset, our tdIF model's power consumption is only about $40\%$ higher than the $16$ time-steps IF-based model.
These results demonstrate that while tdIF models increase energy consumption during neuronal integration and exhibit higher spike firing rates, their overall power consumption remains within the same order of magnitude as IF models while delivering superior performance.

\section{Conclusion}

In this work, we explore the potential of Spiking Neural Networks in visual detection tasks.
This work elucidates the impact of residual membrane potential errors on rate-based method, alongside the influence of quantization errors on the accuracy of regression vision tasks.
Addressing the challenges posed by these errors, our idea is simple and intuitive.
To this end, we employ delay-spike strategy to mitigate residual potential errors arising from irregular spiking, and utilize temporal-dependent Integrate-and-Fire neuron architecture to enhance the feature representation capability of neural networks.
The extensive experimental validation confirms that our proposed method achieves the state-of-the-art performance metrics across multiple vision tasks, delivering superior accuracy while maintaining low latency in both object detection and lane detection.
Furthermore, our proposed neuron model demonstrates power consumption on neuromorphic hardware that remains within the same order of magnitude as conventional IF neurons.
In summary, this research represents a substantial leap forward in deploying SNNs for visual detection applications, establishing a foundational framework for subsequent investigations in neuromorphic vision systems.


\section{Impact Statement}
Our paper proposes the delay-spike mechanism and the temporal-dependent Integrate-and-Fire (tdIF) neuron model, achieving low-latency performance in visual regression tasks.
For any hardware device implementing layer-by-layer computation in spiking neural networks (SNNs), the delay-spike mechanism significantly reduces errors caused by residual membrane potential after conversion, without increasing computational complexity. 
Furthermore, the tdIF model demonstrates a remarkable reduction in latency for regression tasks, suggesting that SNNs can achieve excellent feature representation capabilities even with low time-steps, provided they fully utilize temporal information. 
This breakthrough highlights the potential of SNNs for ultra-low-power applications, paving the way for their widespread adoption in real-time, energy-efficient AI systems.

\nocite{langley00}


\bibliographystyle{icml2025}

\newpage
\appendix
\onecolumn

The experimental results in this paper are reproducible. We explain the details of model training and configuration in the main text and supplement it in the appendix. 
\subsection{Experimental Setting.} \label{apx:es} 
The implementation is in Pytorch and all the experiments are computed on a machine with an NVIDIA A800 GPU.
\subsubsection{Pascal VOC and MS COCO}
The Pascal VOC dataset \cite{everingham2010pascal} contains $16551$ training images and $4952$ validation images. 
For training preprocessing, we employ a range of multiscale image sizes  to randomly resize the images.
The actual multiscale ranges is $[416-r\times32, 416+r\times32]$, where $r \in [0,5]$.
In data augmentation,  the strategies of Mosaic and MixUp from \cite{bochkovskiy2020yolov4} are utilized to enhance the diversity and robustness of training datasets.
In MixUp, four different images are selected and placed in a specific manner (in a $2\times2$ grid) to form a new large image.
We use $ \alpha \times image_1 + (1 - \alpha) \times image_2$ ,where $\alpha=0.5$, to generate our mixing image data.
Additionally, We employ Exponential moving average (EMA) technology  \cite{cai2021exponential} in our training process.
The learning rate is set to $0.01$ and followed by a cosine decay schedule same as \cite{zheng2021yolox}.
The Weight decay is set to $5\times10^{-4}$, and the networks are optimized for 90 epochs.
For test images, they are padded and resized to the same size. 
For all architectures we tested, the Max Pooling layers are replaced to Average Pooling layers.
The input features of the YOLOv3 head are adjusted according to the number of channels in the output feature maps of ResNet34.

The MS COCO dataset \cite{wang2023toward} contains $118k$ training images and $5k$ validation images. 
The training technique remains consistent with our approach for training on VOC, but with modifications to the input range $[640-r\times32, 640+r\times32]$.

\subsubsection{Tusimple and CULane}
Tusimple \cite{pizzati2020lane} is a dataset of highway driving scenes, which contains $3.3k$ training images and $2.8k$ validation images.
CULane\cite{pan2018spatial}. is a widely used large lane detection dataset with 9 different scenarios (Urban\&highway),  which contains $88.9k$ training images and $34.7k$ validation images.

For training preprocessing, we utilize techniques such as Random Blur, Brightness Adjustment, Rotation, and Resizing to augment the images.
In the optimizing process, we use Adam optimizer and step learning rate decay  with an initial learning rate of $3e-4$. 
For each dataset, we train on the training set without any extra data. 
We respectively train $20$ and $16$ epochs for CULane and TuSimple with a batchsize of $128$. 

For CULane, we  utilizes the F1 measure as in \cite{liu2021condlanenet}.
IoU between the predicted lane line and GT label is taken for judging whether a sample is true positive (TP) or false positive (FP) or false negative (FN). 
IoU of two lines is defined as the IoU of their masks with a fixed line width.
F1-measure is calculated as follows:
\begin{align}
        \label{Eq:F1_metric} 
        F1 &= \frac{2 \times Prec \times Recall}{Prec+Recall}, \\  \quad  Prec &= \frac{TP}{TP+FP} \\ Recall &= \frac{TP}{TP+FN}
\end{align}

For TuSimple dataset, there are three official indicators: false-positive rate (FPR), false-negative rate (FNR), and accuracy.
\begin{align}
        \label{Eq:acc_metric} 
        Acc &= \frac{\sum_{clip}C_{clip}}{\sum_{clip}S_{clip}}
\end{align}
Where $C_{clip}$ is the number of correctly predicted lane points and $S_{clip}$ is the total number of lane points of a clip.

\subsection{Additional results.}\label{apx:addres}
Our method employs various quantized time-steps to train the model and obtain the corresponding Spiking Neural Network performance (the same time-steps) based on these model weights. 
Here, we present the progress of conversion under different training models in Table \ref{table:add_res}, serving as supplementary information for further experimental results.
\newpage
\begin{table*}[htbp]
        \caption{\textbf{The trade-off between accuracy of the dataset and inference latency.} 
        }
        \centering
        \tabcolsep=0.3cm     
        \renewcommand{\arraystretch}{0.85}
        \begin{tabular}{l|c|c|c|c|c|c|c|c}
        \hline

        \multirow{2}{*}{Architecture}   & \multirow{2}{*}{\makecell{Quantized\\time-steps}} & \multirow{2}{*}{\makecell{ANN\\metric}}  & \multicolumn{6}{c}{time-steps} \\ \cline{4-9}
                                        &                        &  & 3    & 4     & 5     & 7     & 8  & 10 \\ \hline\hline
        \multicolumn{9}{c}{mAP@0.50 Results on Pascal VOC }    \\ \hline
        \multirow{4}{*}{YOLOv3-Tiny}    &    32                  & 64.30& 60.28 & 63.62 & 63.96 & 64.17 & 64.08 &64.09 \\ 
                                        &    64                  & 64.74& 60.24 & 63.72 & 64.43 & 64.80 & 64.80 &64.81\\ 
                                        &    128                 & 64.43& 59.83 & 63.71 & 64.46 & 64.47 & 64.61 &64.54\\ 
                                        &    256                 & 64.57& 60.34 & 63.77 & 64.19 & 64.60 & 64.56 &64.81\\ \hline
        \multirow{4}{*}{\makecell{YOLOv3\\(ResNet34)}}& 32          & 74.44& 54.01 & 72.46 & 74.42 & 74.80 & 74.64 &74.79\\ 
                                        &    64                  & 77.89& 25.90 & 71.62 & 77.14 & 78.10 & 77.89 &77.96\\ 
                                        &    128                 & 79.01& 11.49 & 66.72 & 77.46 & 79.01 & 78.95 &78.86\\ 
                                        &    256                 & 78.48& 3.73  & 53.16 & 76.12 & 78.39 & 78.55 &78.79\\ \hline\hline
        \multicolumn{9}{c}{mAP@0.50 Results on MS COCO }    \\ \hline
        \multirow{4}{*}{YOLOv3-Tiny}    &    32                  & 36.17& 32.52 & 35.67 & 36.12 & 36.28 & 36.31 &36.19 \\ 
                                        &    64                  & 36.40& 31.84 & 35.46 & 36.23 & 36.48 & 36.52 &36.52\\ 
                                        &    128                 & 36.27& 31.40 & 35.53 & 35.88 & 36.25 & 36.27 &36.22\\ 
                                        &    256                 & 36.31& 31.48 & 35.43 & 36.00 & 36.22 & 36.29 &36.30\\ \hline
        \multirow{4}{*}{\makecell{YOLOv3\\(ResNet34)}}& 32          & 56.21& 41.95 & 53.70 & 55.73 & 56.02 & 56.03 &55.98\\ 
                                        &    64                  & 59.21& 25.79 & 52.50 & 57.08 & 58.24 & 58.06 &58.19\\ 
                                        &    128                 & 60.51& 11.90 & 50.78 & 57.50 & 58.89 & 58.82 &58.85\\ 
                                        &    256                 & 60.11& 9.53  & 49.58 & 58.21 & 59.70 & 59.78 &59.77\\ \hline\hline
        \multicolumn{9}{c}{Acc(\%)  Results on Tusimple }    \\ \hline
        \multirow{4}{*}{\makecell{ResNet18\\+Condlane}}&    16      & 94.82& 45.99 & 94.80 & 94.79 & 94.71 & 94.78 & 94.77\\
                                        &    32                  & 95.34& 19.85 & 89.21 & 95.35 & 95.40 & 95.39 & 95.42\\ 
                                        &    64                  & 95.48& 12.08 & 65.47 & 95.00 & 95.52 & 95.54 & 95.53\\ 
                                        &    128                 & 95.38& 10.62 & 58.53 & 94.65 & 95.37 & 95.39 & 95.37\\  \hline 
        \multirow{4}{*}{\makecell{ResNet34\\+Condlane}}& 16         & 93.62& 14.26 & 93.55 & 87.45 & 69.67 & 68.54 & 68.32\\ 
                                        &    32                  & 95.53& 0.45  & 85.83 & 95.48 & 95.71 & 95.72 & 95.71\\ 
                                        &    64                  & 95.65& 0.41  & 61.82 & 95.31 & 95.68 & 95.71 & 95.70\\ 
                                        &    128                 & 95.53& 0.79  & 54.90 & 94.50 & 95.53 & 95.54 & 95.58\\ \hline
        \multicolumn{9}{c}{F1(\%)  Results on CULane }    \\ \hline
        \multirow{4}{*}{\makecell{ResNet18\\+Condlane}}&    16      & 69.41& 35.69 & 69.32 & 69.80 & 70.03 & 70.04 & 70.03\\
                                        &    32                  & 72.13& 26.90 & 65.01 & 72.18 & 72.81 & 72.81 & 72.91\\ 
                                        &    64                  & 72.26&  8.00 & 59.65 & 70.84 & 72.60 & 72.60 & 72.62\\ 
                                        &    128                 & 72.22&  3.26 & 48.61 & 68.90 & 72.24 & 72.29 & 72.30\\  \hline 
        \multirow{4}{*}{\makecell{ResNet34\\+Condlane}}& 16         & 65.33&  8.73 & 65.37 & 63.34 & 58.21 & 58.29 & 58.12\\ 
                                        &    32                  & 76.68&  3.45 & 67.33 & 76.64 & 77.73 & 77.74 & 77.77\\ 
                                        &    64                  & 77.02&  0.53 & 64.05 & 75.89 & 77.25 & 77.39 & 77.38\\ 
                                        &    128                 & 77.59&  0.12 & 49.86 & 74.28 & 77.16 & 77.27 & 77.31\\ \hline
        \end{tabular}
        \label{table:add_res}
\end{table*}

\end{document}